\documentclass{article}

\usepackage{PRIMEarxiv}

\usepackage[utf8]{inputenc} 
\usepackage[T1]{fontenc}    
\usepackage{hyperref}       
\usepackage{url}            
\usepackage{booktabs}       
\usepackage{amsfonts}       
\usepackage{nicefrac}       
\usepackage{microtype}      
\usepackage{lipsum}
\usepackage{fancyhdr}       
\usepackage{graphicx}       
\graphicspath{{media/}}     
\usepackage{authblk}       
\usepackage{csquotes}
\usepackage{amsmath}
\usepackage{amsfonts}
\usepackage{algorithm}
\usepackage{algorithmic}
\usepackage{authblk}
\usepackage{fancyhdr}
\usepackage{amsmath}
\usepackage{graphicx}
\usepackage[numbers,square,sort&compress]{natbib}

\usepackage[english]{babel}
\usepackage{amsmath}
\usepackage{graphicx}

\title{Distilling and exploiting quantitative insights from Large Language Models for enhanced Bayesian optimization of chemical reactions
}

\author[1]{Roshan Patel}
\author[2]{Saeed Moayedpour}
\author[1]{Louis De Lescure}
\author[2]{Lorenzo Kogler-Anele}
\author[1]{Alan Cherney}
\author[2]{Sven Jager \thanks{Corresponding author: \text{Sven.Jager@sanofi.com}} $^{}$}
\author[1]{Yasser Jangjou \thanks{Corresponding author: \text{Yasser.Jangjou@sanofi.com}} $^{}$}

\affil[1]{CMC Synthetics Platform, Sanofi, 350 Water St, Cambridge, MA, 02141, USA
}
\affil[2]{Digital R\&D, Sanofi, 450 Water St, Cambridge, MA, 02141, USA
}

\begin{document}
\maketitle

\begin{abstract}
Machine learning and Bayesian optimization (BO) algorithms can significantly accelerate the optimization of chemical reactions. Transfer learning can bolster the effectiveness of BO algorithms in low-data regimes by leveraging pre-existing chemical information or data outside the direct optimization task (i.e., source data). Large language models (LLMs) have demonstrated that chemical information present in foundation training data can give them utility for processing chemical data. Furthermore, they can be augmented with and help synthesize potentially multiple modalities of source chemical data germane to the optimization task. In this work, we examine how chemical information from LLMs can be elicited and used for transfer learning to accelerate the BO of reaction conditions to maximize yield. Specifically, we show that a survey-like prompting scheme and preference learning can be used to infer a utility function which models prior chemical information embedded in LLMs over a chemical parameter space; we find that the utility function shows modest correlation to true experimental measurements (yield) over the parameter space despite operating in a zero-shot setting. Furthermore, we show that the utility function can be leveraged to focus BO efforts in promising regions of the parameter space, improving the yield of the initial BO query and enhancing optimization in 4 of the 6 datasets studied. Overall, we view this work as a step towards bridging the gap between the chemistry knowledge embedded in LLMs and the capabilities of principled BO methods to accelerate reaction optimization.
\end{abstract}


\section{Introduction}

Machine learning and data-driven approaches can significantly accelerate the optimization of chemical processes \cite{golem,rlchemopt,natchemopt,mlcompaid}. In applications where data is insufficient for comprehensive predictive modeling (e.g., high-throughput screening), Bayesian optimization (BO) algorithms stand out as data-efficient methods to iteratively navigate the chemical and process parameter space to target desired properties from the chemical product \cite{humanoutofloop,shieldsbo,bochempro,chemicalbo}. For example, Shields et al. \cite{shieldsbo} show that BO can work well to identify chemicals (e.g., base, solvent, catalyst ligands) and reaction conditions (e.g., temperature, chemical concentration) to maximize the yields of Buckwald-Hartwig coupling, Suzuki- Miyaura coupling, and direct arylation reactions. We refer readers to a recent review by Guo and Rankovic et al.\cite{boreview} for comprehensive discussion on successful applications of BO for chemical process development. 

Transfer learning can significantly accelerate BO-lead workflows by leveraging (source) information or data outside of the direct domain of a given optimization task \cite{tlbo,tl2,tlboreview}. For example, source datasets can be used to better inform model development for the domain task \cite{tlmodel,tlmodel1,tlmodel2}. In addition, source data can be used to identify and focus optimization efforts on promising regions of the parameter space through modification of the acquisition function \cite{tlbo,tlbo1,tlbo3,tlaqf,pibo,prioropt,mtbo}. Overall, though, the application of these and other transfer learning strategies for BO heavily rely on the identification, curation, and numerical encoding of relevant source datasets which are often difficult and laborious to accomplish in practice. Furthermore, qualitative information outside organized datasets (e.g., insights / conclusions found as text in research articles) are typically not leveraged for transfer learning despite representing a large volume of information pertinent for new chemical design tasks. 

In recent years, large language models (LLMs) have demonstrated that its ability to model natural language can help perform challenging tasks in disparate chemical domains \cite{llmchem,llmchem2}. For example, with in-context learning, LLMs have been used as regression and classification models to predict chemical properties \cite{llmchem1,llmbo1,llmregress,llmreggpt3}. In addition, LLMs have shown promise for designing experiments in the context of chemical optimization problems or making meta decisions about the optimization process (e.g., integrating several systems / software needed to execute optimization or deciding a suitable stopping condition) \cite{chemx,e2e,chemcrow,llmbo1,rankovi,sober}. Given these observations, we posit that LLMs can be queried to transfer pertinent chemical information from source data (e.g., foundation model training data, fine-tuning data) to target Bayesian optimization campaigns and accelerate process development. 

In this study, we examine how information from LLMs can be distilled and used to accelerate Bayesian reaction optimization through transfer learning. Specifically, we show that preference learning \cite{preference_summary,Chu2005PreferenceLW} can be used to infer a utility function over the reaction parameter space from LLM-answered surveys that shows modest correlation to measured reaction yields; promisingly, we accomplish this despite operating in a zero-shot setting with no in-context learning \cite{llmbo1,llmreggpt3,llmregress,icl_review} or fine-tuning \cite{llmreggpt3,sober,fine_tune_summary}. Furthermore, we show that when incorporated in the acquisition function, the utility function can be used to focus BO queries in promising regions of the parameter space. We observe that this significantly improves the yield of the initial query to seed BO and enhances optimization in 4 of the 6 datasets studied. Overall, we view this work as a step towards bridging the gap between the chemistry knowledge embedded in LLMs and the capabilities of principled BO methods to accelerate reaction optimization.

\section{Methods}

\subsection{Datasets}
We explore our approach with six chemical reaction datasets compiled by Shields et al. Datasets 1-5 correspond to Buchwald-Hartwig (BH) reactions and each contain 792 recorded experiments. Experiments in these datasets are characterized by four reaction parameters: the identity of a specific aryl halide reactant, the palladium precatalyst, the additive, and the base used for the reaction and are labeled by a measured product yield. Dataset 6 corresponds to a direct arylation (DA) reaction and contains 1728 experiments. Experiments in this dataset are characterized by five reaction parameters: the identity of a palladium catalyst ligand, the base, the solvent, temperature, and concentration and are also labeled with a measured product yield. The objective for all datasets is to identify the experiment, characterized by a specific set of reaction parameters, that will give the maximum product yield.

\subsection{Formulation and implementation of approach}\label{S:method}
\begin{figure}[t]
\centering
\includegraphics[width=5.5in,keepaspectratio=True]{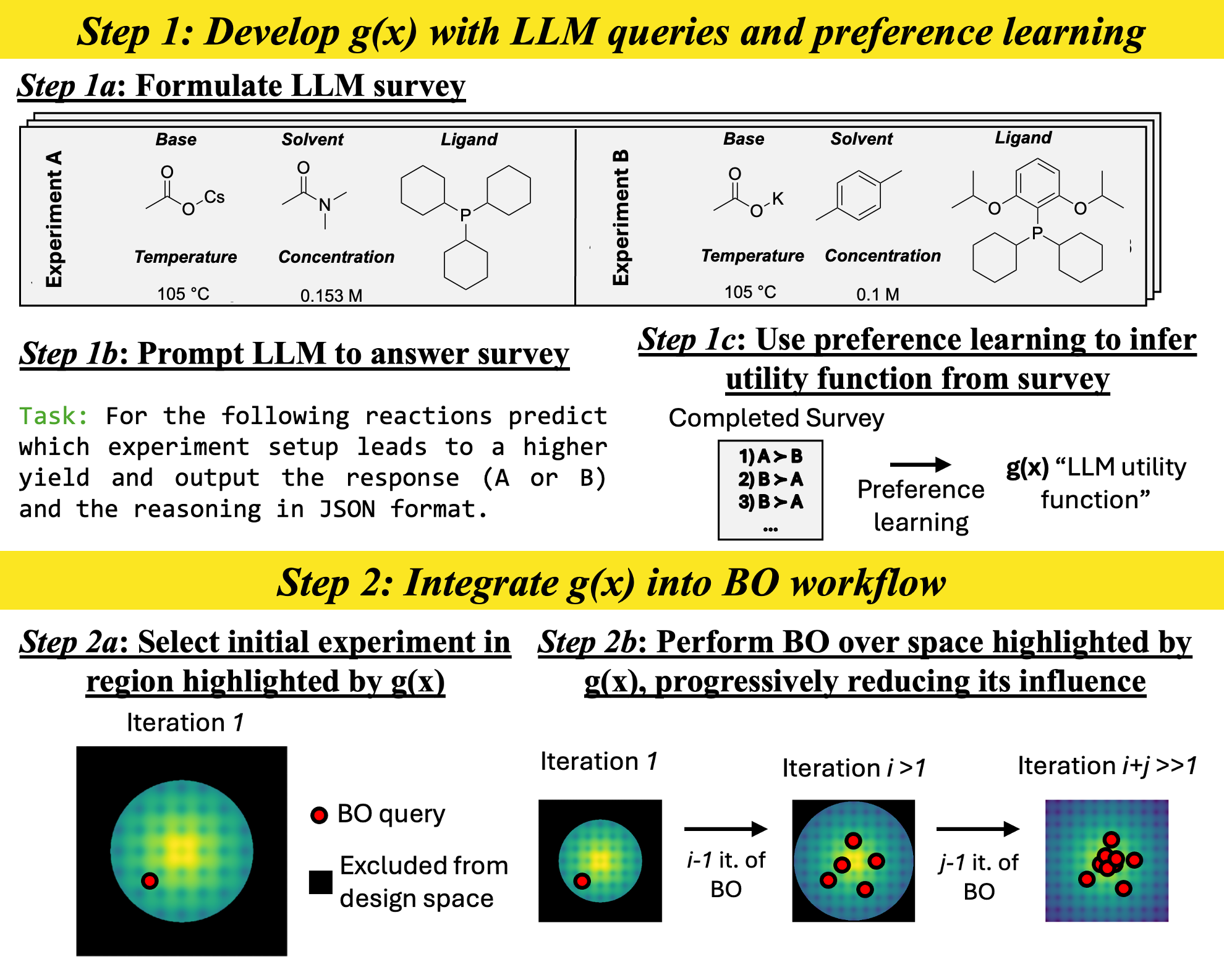}
\caption{A schematic outlining the major elements of the approach in this study.  \label{fig:skem} }
\end{figure}

The overall approach taken for each dataset is qualitatively presented in Figure \ref{fig:skem}. Step 1 of the approach aims at distilling chemical insights from the LLM in the form of a utility function $g(x)$. In step 1a, we formulate a survey in which each question presents two experiments, each characterized by a different set of experimental parameters. In step 1b, we prompt the LLM to answer the survey, selecting which experiment (A or B) it predicts will give the higher yield for each question. In step 1c, we use preference learning to infer a utility function $g(x)$ based on the preferences expressed as choices in the survey. Since we prompt the LLM to prefer experiments with higher predicted yields, we expect $g(x)$ to correlate to yield and thus represent useful, quantitative prior information from the LLM over the set of experiments in a dataset. 

Step 2 of the approach aims at leveraging $g(x)$ to expedite BO of reaction parameters. As discussed more in section \ref{S:pbo}, we use $g(x)$ to identify promising regions of the parameter space and constrain optimization to these experiments only. Thus in step 2a, we begin optimization by randomly selecting an experiment in the promising set of experiments. In step 2b, we perform BO, gradually removing restrictions on the design space imposed by $g(x)$ as more data is available for surrogate modeling. The pseudo code for the approach is provided at the end of section \ref{S:pbo}.

\subsubsection{Chemical reaction parameter optimization with Bayesian optimization}
Reaction parameter optimization in this work is viewed as a black-box optimization problem: 
\[
x^{*} = \operatorname*{argmax}_{x\in X} f(x)
\]
where $x$ is a variable that represents a single candidate experiment, $X$ represents the full set of candidate experiments considered as possible designs, and $f(x_i)$ represents the noiseless output quantity (e.g., yield) that should result from performing experiment $x_i$; typically, we can access noisy measurements of $f(x_i)$ through experiments: $y_i$ = $f(x_i)$ + $\epsilon$. In our work, we represent experiments $x$ as a concatenation of one-hot-encoded categorical variables (e.g., base, ligand, solvent identity) and continuous variables (e.g., concentration and temperature). Given that obtaining measurements can be time consuming / expensive, the goal is to identify the $x^*$ among all candidate experiments that gives the maximum value of $f(x)$ with as few experiments as possible.
Bayesian optimization (BO) is an iterative approach utilizing probabilistic modeling that can be used to solve this class of optimization tasks. At iteration $n$ of BO, we have measured the output of $n$ experiments giving us dataset $D_n = \{(x_i, y_i)\}_{i=1}^{n}$. Following, the dataset is used to develop a surrogate model $\hat{f}$ used to predict the output of a given experiment and estimate an uncertainty in that prediction. Gaussian process regression models \cite{Rasmussen2004} and Bayesian neural networks \cite{Jospin_2022} are both common surrogate modeling strategies, offering principled methods to estimate a predictive posterior distribution $p(y\hat{}_i \mid D_n, x_i)$. We use the modeling strategy developed by Shields et al. \cite{shieldsbo}, which leverages GPR with specific priors on kernel parameters suitable for the experiment representation strategy described in this work. The surrogate model is used to compute terms in an acquisition function $\alpha(x, D_n)$, the maximizing argument of which is selected as the next best experiment to obtain a measurement for: 
\[
x^{**} = \operatorname*{argmax}_{x\in X} \alpha(x, D_n)
\]
The expected improvement (EI) function is a popular choice among studied acquisition functions and has been applied extensively for chemical design \cite{humanoutofloop,benchmark}:
\begin{equation}
\alpha(x, D_n) = \mathbb{E}_{y\sim p(y \mid D_n, x)}[\text{max}(y_{\text{max},n}-y,0)]
\end{equation}
where $y_{\text{max},n}$ is the largest measurement $y$ found in $D_n$. We use this acquisition as the baseline for our work. Once the maximizing argument $x^{**}$ is found and the measurement $y^{**}$ for the corresponding experiment is taken, $x^{**}$ and $y^{**}$ are added to the dataset (($x^{**}$,$y^{**}$) $\cup$ $\{(x_i, y_i)\}_{i=1}^{n})$ and the next iteration of BO begins. Typically, optimization efforts are concluded when a budget for iterative experimentation has been depleted or improvement upon the largest observed measurement has stagnated for several iterations.

\subsubsection{Extracting chemical insights from LLMs}

In this section, we discuss our approach to extract quantitative, chemical insights from LLMs in the form of a utility function $g(x)$. First, for each dataset, we formulate a survey consisting of several questions. For each question in a survey, the LLM is presented with two experiments from the dataset (characterized by reaction parameters) and is subsequently prompted to select which of the two would result in a higher yield and to provide its reasoning. An example of a question prompt and the LLM response is provided in Figure S2 of the supporting information. To design the survey questions $S_{\text{unanswered}}$ for a given dataset, we created two identical arrays, where each array contains $L$ instances of every experiment in a dataset. Then, we randomly paired elements between each array to form questions, removing repeated questions and questions where the paired experiments were identical. For datasets BH1-BH5 we set $L$ to 10 and for DA $L$ was set to 5 (to keep the total number of questions roughly similar to the BH surveys). This resulted in 7792,7842,7788,7825,7804, and 8610 survey questions designed for the BH1-5 and DA datasets respectively. Overall, we hypothesized that surveys generated using this procedure would facilitate expressing hierarchical preferences over the full set of experiments and subsequent preference learning. The LLM model used in this study was Claude 3.5 Sonnet.

A completed survey is represented as $S_{\text{answered}} = \{(x_{i,j} \succ x_{i,k})\}_{i=1}^{m}$ where experiment $x_{j}$ is preferred over experiment $x_{k}$ in question $i$ of the survey with $m$ total questions. Following, we leverage preference learning to infer a utility function $g(x)$ aligns with LLM predictions made in the survey: namely, $g(x_{j})$ should be greater than $g(x_{k})$ if $x_{j} \succ x_{k}$. Since the LLM was prompted to prefer experiments with higher expected yield based on its chemical reasoning, we expect $g(x)$ to correlate to the true experimental yield measured for experiments in a dataset. We follow the approach of Chu and Gharamani \cite{Chu2005PreferenceLW}, who model $g(x)$ as a Gaussian process and define a function to model the likelihood of observing a preference among pairs of options given their values from the utility function (assumed to contain noise). To tune hyperparameters (e.g., parameters of the kernel), they use the Laplace approximation to define an expression of the posterior density over utility functions conditioned on the data and optimize it (MAP estimate). For our work, we employ the Botorch \cite{botorch} implementation of Chu and Gharamani's approach using the PairwiseGP module. Upon training, we take the mean of the GP posterior conditioned on the survey data as the utility function $g(x)$ and to be a representation of prior chemical knowledge embedded in the LLM over the chemical parameter space. 

\subsubsection{Leveraging chemical insights from LLMs for enhanced optimization}\label{S:pbo}

A common way to incorporate prior-knowledge or information in the BO algorithm is through an adjustment of the acquisition function. For example, Souza et al.\cite{prioropt} and Hvarfner et al. \cite{pibo}. weight the standard BO acquisition function with a decaying (as a function BO iterations) prior probability function that computes the probability $\pi$ that experiment $x$ maps to the maximum of $f$. In doing so, the acquisition function is biased to explore promising regions of the parameter space encoded in $\pi(x)$ in early iterations of BO. Our work follows their weighting framework, computing the modified acquisition function as: 
\begin{equation}
\alpha_{\pi,n}(x, D_n,n) = \alpha(x, D_n) \pi(g(x),p(n))
\end{equation}
$\pi$ is computed as a simple indicator function: 
\[
\pi(g(x),p(n)) =
\begin{cases}
1 & \text{if } g(x) \geq P_{p(n)}  \\
0 & \text{if } g(x) < P_{p(n)}
\end{cases}
\]
where $P_p$ is $p$th percentile value of the set $G = \{g(x) \mid x \in X\}$. This binary weighting to the acquisition allows optimization to focus on promising regions of the chemical space highlighted by $g$, without further biasing candidate selection with potentially noisy utility values. Our approach can also be viewed as design space pruning \cite{dsp,dsp1,dsp2}, where unpromising portions of the design space are excluded from the set $X$ of candidate experiments. Given that our weighting / pruning approach may adversely impact optimization if $g$ is negatively correlated with $f$ (or by excluding the true maximizing argument of $f$), we recommend setting percentile $p(n)$ as a decaying function of BO iterations $n$ such that $p\rightarrow0$ as $n\rightarrow\infty$. In effect, this relaxes the constraint on the design space imposed by $g$ for candidate selection as more experiments are performed and the surrogate model $\hat{f}$ becomes increasingly reliable. In our work, we select $p(n)$ as a simple 2-step function; see Figure S1 for additional details on how we design step function $p(n)$ for our work.

\begin{algorithm}
    \caption{Psuedo-code for LLM-augmented BO}\label{your_label}
    \begin{algorithmic}

        \STATE  \textbf{Input}: Parameter space $X$, Number of BO queries $N$, Acquisition function $a$, Percentile function $p(n)$, Experiment instances $L$

        \STATE \textbf{Step 1: Develop utility function $g(x)$ via LLM queries and preference learning}

        \STATE Pair elements of two identical arrays containing $L$ instances of each experiment to formulate the survey $S_{\text{unanswered}} = \{(x_{i,j}, x_{i,k})\}_{i=1}^{m}$, $x_{i,j},x_{j,k} \in X$

        \STATE For each question in the survey prompt the LLM to predict which experiment will give a higher yield: $S_{\text{answered}} = \{(x_{i,j} \succ x_{i,k})\}_{i=1}^{m}$

        \STATE Use preference learning to infer utility function: $g(x) \xleftarrow{}\text{Train(} S_{\text{answered}},X)$

        \STATE \textbf{Step 2: Integrate utility function $g(x)$ into BO workflow}
        
        \STATE Evaluate $g(x)$ over the parameter space: $G = \{g(x) | x\in X\}$
        
        \STATE Randomly select an experiment from the set of promising experiments: $x_0 \sim \{x | \pi(g(x),p(n=0)) = 1  \text{ and } x\in X\}$
        
        \STATE Initialize dataset for BO with this point and its measured label: $D_0=\{(x_0,y_0)\}$

        \FOR{n = $1$ to $N-1$}

            \STATE Train surrogate model: $\hat{f}(x)\leftarrow \text{Train}(D_n) $
            \STATE Obtain candidate by optimizing acquisition function: $x^{**} = \operatorname*{argmax}_{x\in X} \alpha(x, D_n) \pi(g(x),p(n))$
            \STATE Augment dataset with this point and its measured label: $D_n=$(($x^{**}$,$y^{**}$) $\cup$ $\{(x_i, y_i)\}_{i=0}^{n-1})$
            
        \ENDFOR

        \STATE \textbf{return} $x^{*} = \operatorname*{argmax}_{(x_i,y_i)\in D_{N-1}} y_i$

    \end{algorithmic}
\end{algorithm}

\section{Related Works}

\subsection{Transfer learning}

In one paradigm of transfer learning, prior information is leveraged to make judicious modifications to the acquisition function to accelerate Bayesian optimization in a target domain \cite{tlbo,rltlbo,aqfbo,tlaqf}. For example, Souza et al.\cite{prioropt} and Hvarfner et al. \cite{pibo} weight the acquisition function with a prior $\pi(x)$ on the function's maxima, biasing the BO algorithm to focus early optimization efforts on regions of the parameter space with high probability mass. As mentioned, our approach of leveraging information encoded in $g(x)$ follows a similar framework. In their work, however, $\pi(x)$ is typically encoded as a parameterized probability function; choosing the type of type of function or specific parameter values to match source data or information can be non-trivial. Along with formulating a new acquisition function to incorporate $\pi(x)$, Adachi et al. \cite{tlbo3} propose using preference learning to distill insights from human experts and obtain $\pi(x)$. Our own experiments suggested that the quality and quantity of data collected in surveys completed by human experts was not sufficient to apply this method for our domain application. We posited that LLMs offer a promising alternative to human experts: they can answer orders of magnitude more questions in a fraction of the time and could leverage chemical information in source data to accurately answer questions. 

\subsection{LLM-augmented Bayesian optimization}

A few recent works have explored how LLMs can be used to accelerate BO in chemical systems; predominantly, these works have leveraged LLMs to inform the development of the surrogate model. One strategy is to use the LLM as the surrogate model itself. For example, Ramos et al. \cite{llmbo1} show that in-context learning and specific prompting strategies (and interpretation of token probabilities) could be used to develop a regressor capable of uncertainty quantification, which they then use for BO. Another approach is to use the LLM to process some description of the chemical system / experiment and produce an embedding from which a surrogate model can be trained to make a prediction. For example, Rankovi{\'c} and Schwaller \cite{rankovi} show that these LLM embeddings are competitive with (and can outperform) those obtained from more sophisticated and domain-informed pre-training procedures. Kristiadi et al. \cite{sober} show that the performance of this approach can be further improved when using domain specific and fine-tuned LLMs. Furthermore, they show that parameter-efficient fine-tuning and Bayesian neural networks can offer a principled way to use the LLM as a surrogate model and allow it to further learn informative embeddings of reactions. Overall, these are promising developments in leveraging source information in LLMs to expedite BO in a target domain. Our work differs from these approaches in that we separate the modeling of the target information (GPR surrogate model $\hat{f}(x)$) and the source information (LLM-derived utility function $g(x)$); instead the source information is included at the point of defining the acquisition function. Overall, the binary weighting scheme we use to adjust the acquisition function accomplishes a similar purpose to what is presented by Liu et al. \cite{llmbohp}, who use the LLM to first pre-select which points are considered for initialization and optimization at a given iteration. We suggest that obtaining and exploiting quantitative information present in the utility function can provide finer control for experiment selection strategies.


\section{Results and Discussion}
\subsection{Survey grades and preference learning outcomes}
\begin{table}[h]\label{T:accuracy}
    \centering
    \begin{tabular}{|c|c|c|c|c|c|c|}
        \hline
        & BH1& BH2& BH3& BH4& BH5 &DA\\ \hline
        LLM Survey Accuracy& 64.0\%& 73.1\%& 70.5\%& 55.9\%& 61.5\%&70.5\%\\\hline
    \end{tabular}
    \caption{LLM response accuracy to the survey questions for BH1-5 and DA datasets. Accuracy was calculated as the percentage of correctly predicted answers in a survey.}
\end{table}
\begin{figure}[t]
\includegraphics[width=6in,keepaspectratio=True]{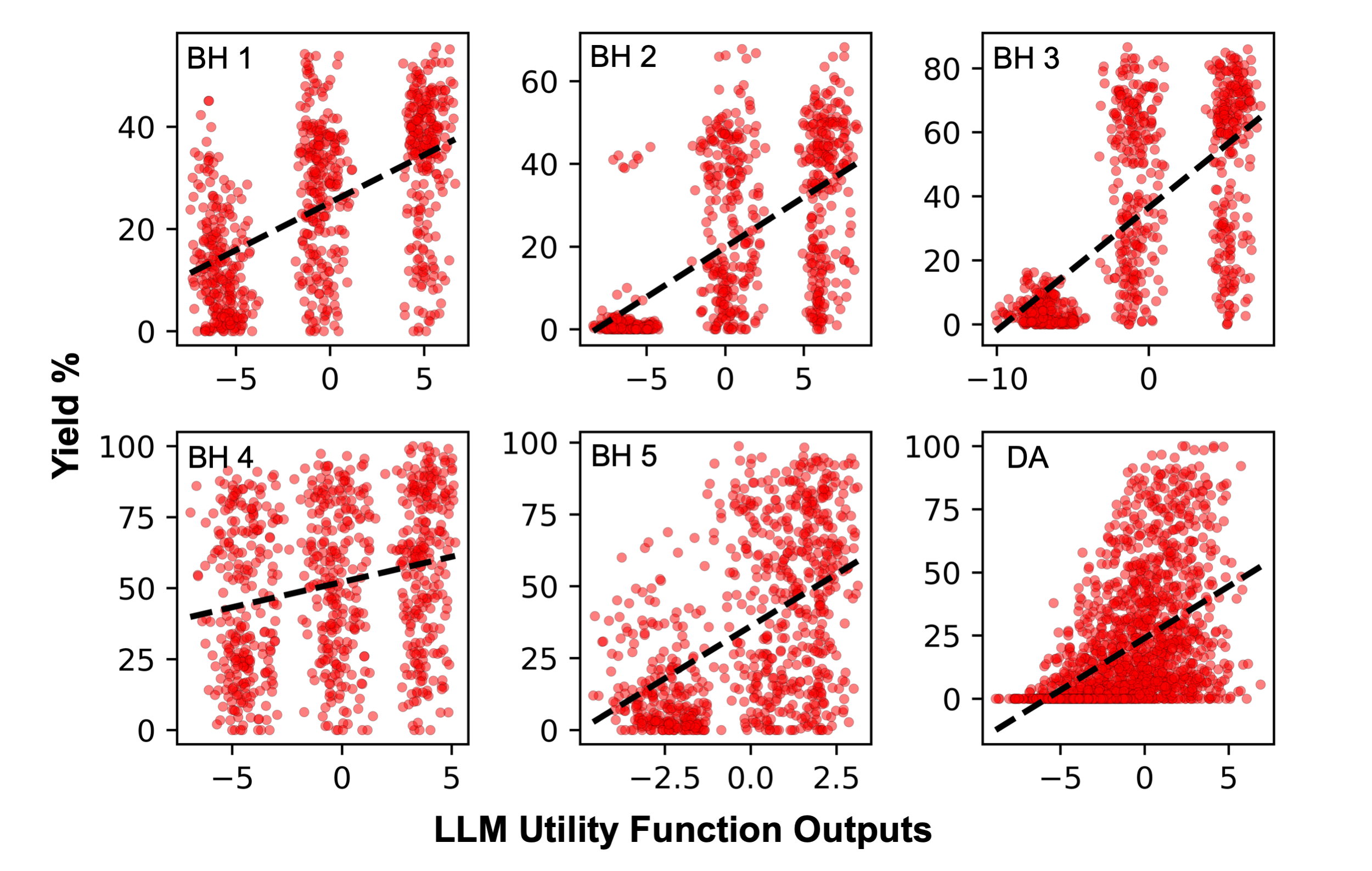}
\caption{\label{fig:corr} Assessing the correlation between utility function outputs computed for experiments across all datasets and their true measured yield. The Pearson r correlation between utility values and yields are 0.55, 0.63, 0.67, 0.22, 0.49, and 0.48 for datasets BH 1, BH 2, BH 3, BH 4, BH 5, and DA respectively, with a p-value $<$ 1e-10 across all datasets. The least squares regression line between utility values and yields is plotted for each panel in a dotted black line to guide the eye.}
\end{figure}

The LLMs were prompted to answer the surveys designed for each dataset and subsequently each survey was graded for accuracy. Specifically, for each question in a given survey, the question was marked ``correct" if the LLM preference (its prediction for whether experiment A or experiment B has the higher yield) aligned with the ground truth; the percentage of questions answered correctly in a given survey is defined as the accuracy. The LLM's accuracy for each survey is reported in Table 1. Overall, we observe that accuracies for all surveys exceed 50\%, suggesting that the LLM could use chemical reasoning present in the foundation model training data to make informed decisions about which of two experiments would result in a higher yield. Figure S2 gives an example of the typical reasoning provided by the LLM in answering survey questions; we observe that decisions are made from relatively simple chemical reasonings (e.g., polarity of the solvent, strength of base, stereochemistry of ligand). Overall though, we find that this is enough to achieve survey accuracies above 50\% (one-tailed binomial test statistically significant for all surveys, p $<$ 0.001), suggesting that despite the simplicity, the chemical information embedded in the LLM is pertinent enough to help make (on average) informed decisions.

Following, for a given dataset, we use the LLM-completed survey and preference modeling to infer the utility function; we compare its output for experiments to their true measured yields. Overall, we observe a positive correlation between utility function outputs and true experimental yields for each dataset (Figure \ref{fig:corr}), indicating that preference modeling could be used to infer a utility function that aligned with the chemically informed, LLM-completed surveys. Importantly, the outputs are not on the same scale as measured yield given that they only encode the utility of a given experiment and not the yield directly. We compared our approach to directly asking the LLM to predict the yield from descriptions of the reaction parameters and for the given reaction (i.e., zero-shot regression), which we observe gives output values that do not correlate positively to yield (Figure S3). Overall, this suggests that the LLM survey + preference modeling approach presented herein is a promising way to distill quantitative insights from LLMs in the zero-shot setting.

Interestingly, for several of the datasets we observe distinct clusters where the utility function gives similar output values for different experiments. For datasets BH 1-4 we observe three distinct clusters, dataset BH 5 has two loosely defined clusters, and dataset DA shows no clustering. Notably, while the mean yield of experiments increases with the mean utility value of each cluster (giving rise to the overall correlation), the yield of experiments within a given cluster correlate relatively poorly to corresponding preference model outputs. This observation motivated the formulation of the approach detailed in section \ref{S:pbo}, where we essentially attempt to restrict BO to query experiments found in clusters with the highest mean preference value and forgo the precise value due to the apparent noise within the cluster.

\subsection{BO experiments}

\begin{figure}[t]
\includegraphics[width=6in,keepaspectratio=True]{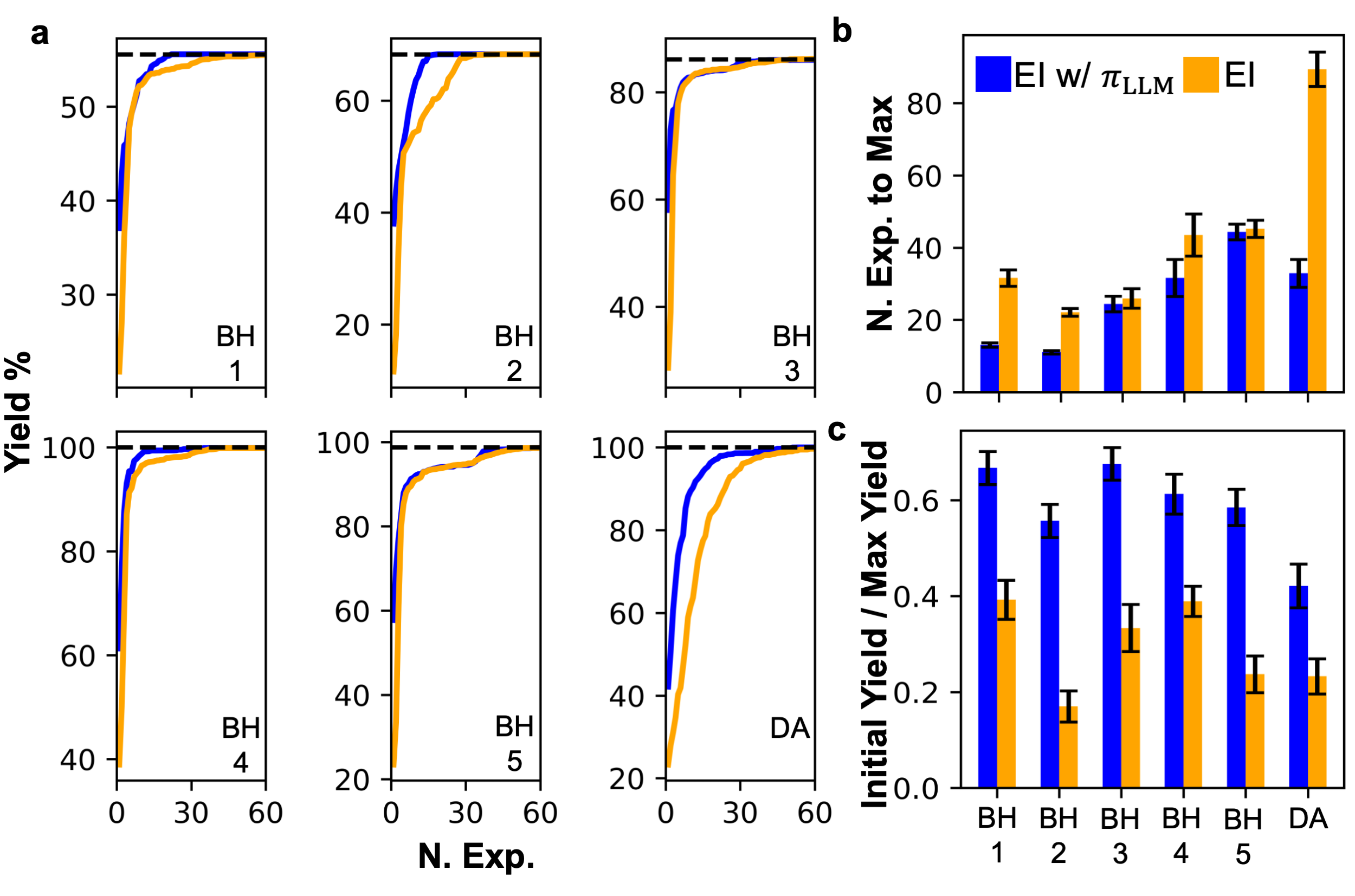}
\caption{\label{fig:BO} Comparing BO of reaction parameters using the expected improvement acquisition function versus the LLM-preference guided expected improvement acquisition. Panel \textbf{a} shows plots the best measured yield observed as a function of the number of experiments performed for each dataset for BO campaigns employing a given acquisition function. The plotted line represents the mean value seen at a given N. exp over the course of n = 50 randomly seeded campaigns. Panel \textbf{b} shows the mean (and standard error from n=50 trials) number of experiments needed to run to observe the maximum yield for a given dataset and acquisition function. Panel \textbf{c} shows the average yield (and standard error from n=50 trials) observed in the initial experiment selected during BO. Values have been normalized by the maximum observed yield for a given dataset.}
\end{figure}

Next, we compare the performance of the expected improvement (EI) acquisition function (eq. 1) to the LLM utility function-modified EI acquisition function (LLM-EI) (eq. 2) for Bayesian reaction parameter optimization using our six datasets. We perform 50 optimization runs for each dataset and acquisition function. Optimization runs for the EI acquisition function are seeded by randomly selecting one experiment out of a given dataset. Optimization runs for the LLM-EI acquisition function are seeded by randomly selecting an experiment among the set $\{x | \pi(g(x),p(n=0)) = 1  \text{ and } x\in X\}$ within a given dataset. For each run, we track the maximum yield observed at a given iteration of BO.

Overall, we observe that the LLM-EI acquisition function significantly outperforms the EI acquisition function for datasets BH1, BH2, BH4, and DA and performs as well as the EI acquisition function for datasets BH3 and BH5 (Figure \ref{fig:BO}a). Specifically, Figure \ref{fig:BO}b shows that the mean number of experiments needed to identify the reaction parameters that give the maximum yield significantly decreases from 32 to 13 (59\% decrease), 22 to 11 (50\% decrease), and 89 to 33 (63\% decrease) for datasets BH1, BH2, and DA respectively when using LLM-EI over EI. Due to difficulties in convergence, the mean number of experiments needed to reach the maximum for each acquisition function show no significant differences for BH4. However, Figure S4 shows that the mean number of experiments needed to find reaction parameters that give 99\% of the maximum attainable yield significantly decreases from 19 to 9 (53\%) for dataset BH4; datasets BH1, BH2, and DA show similar levels of improvement. Furthermore, Figure \ref{fig:BO}c shows that BO-seeding experiments chosen using the utility function on average have significantly higher yields than those chosen completely at random, for all datasets. This may be helpful in applications where a modest yield or property value is needed to progress development, as opposed to near-maximum values. Overall, these results suggest that outputs from utility functions inferred from LLM-completed surveys can help identify promising regions of the chemical space and facilitate BO of chemical parameters.

\section{Discussion and Conclusion}

In this study, we presented an approach to distill and use quantitative insights from LLMs to accelerate Bayesian reaction optimization with transfer learning. Specifically, we prompted the LLM to complete surveys in which each question of a survey asks the LLM to predict which of two experiments is expected to provide the higher yield. We find that the LLM typically employs simple chemical logic to make predictions which led to (on average) correct predictions in surveys. Following, for each dataset, we used preference learning to infer a utility function $g(x)$ which quantitatively models LLM preferences expressed in a survey. We found that the outputs of utility functions show modest correlation to the true yield measured for experiments in a given dataset; thus we interpret $g(x)$ as an expression of prior information provided by the LLM over the chemical parameter space. Lastly, we show that the outputs of $g(x)$ can be used to focus BO queries on promising regions of the parameter space, leading to significantly enhanced optimization in 4 out of the 6 datasets examined and higher experimental yields for initial BO queries. Moving forward, we anticipate several avenues of exploration to improve these results. In ongoing work, we are considering fine-tuning the LLM or using in-context learning with domain-specific literature (possibly identified via document retrieval systems) to further improve the LLM's performance on surveys, which could enhance the correlation between $g(x)$ and yield, enabling BO to better pinpoint promising regions of the design space. Additionally, we are investigating the sensitivity of our approach to several hyperparameters and choices made in this study, such as the number and selection of survey questions, prompt formulation, and the integration of $g(x)$ into the acquisition function.

\section{Supporting Information}
\setcounter{figure}{0}

For our work, we constructed p(n) as a step function parameterized by four values:

\[
p(n) =
\begin{cases}
v_1 & \text{if } n \leq c_1 \\
v_2 & \text{if } c_1<n \leq c_2
\end{cases}
\]
To optimize these four parameters, we perform 100 trials of random search over the parameter space: $v_1 \in(50,100)$, $v_2 \in (0,v1)$, $c_1 \in (0,100)$, $c_2 \in (c1,100)$ and select the parameter set that gives the best average performance among datasets BH1-5. The performance $a$ of a particular parameter set for a given dataset evaluated as:

\[
a = \frac{1}{50}\sum_{i=1}^{50}\sum_{j=1}^{100} y_{ij}
\]
where $y_{ij}$ is the maximum yield observed among experiments observed at iteration $j$ of BO trial $i$ employing the LMM-EI acquisition function. Optimization yields parameters $v_1 = 85\%$, $v_2=15\%$, $c_1 = 30$, $c_2=40$, giving the $p(n)$ shown in Figure S1. We use this $p(n)$ for all BO trials employing the LLM-EI acquisition function in Figure \ref{fig:BO}. The good performance of LLM-EI on the DA dataset shown in Figure \ref{fig:BO} (not used for parameter optimization) serves as validation of the $p(n)$ designed in our study. 

\begin{figure}[htb]
\renewcommand\figurename{Fig. S}

\centering
\includegraphics[width=3in,keepaspectratio=True]{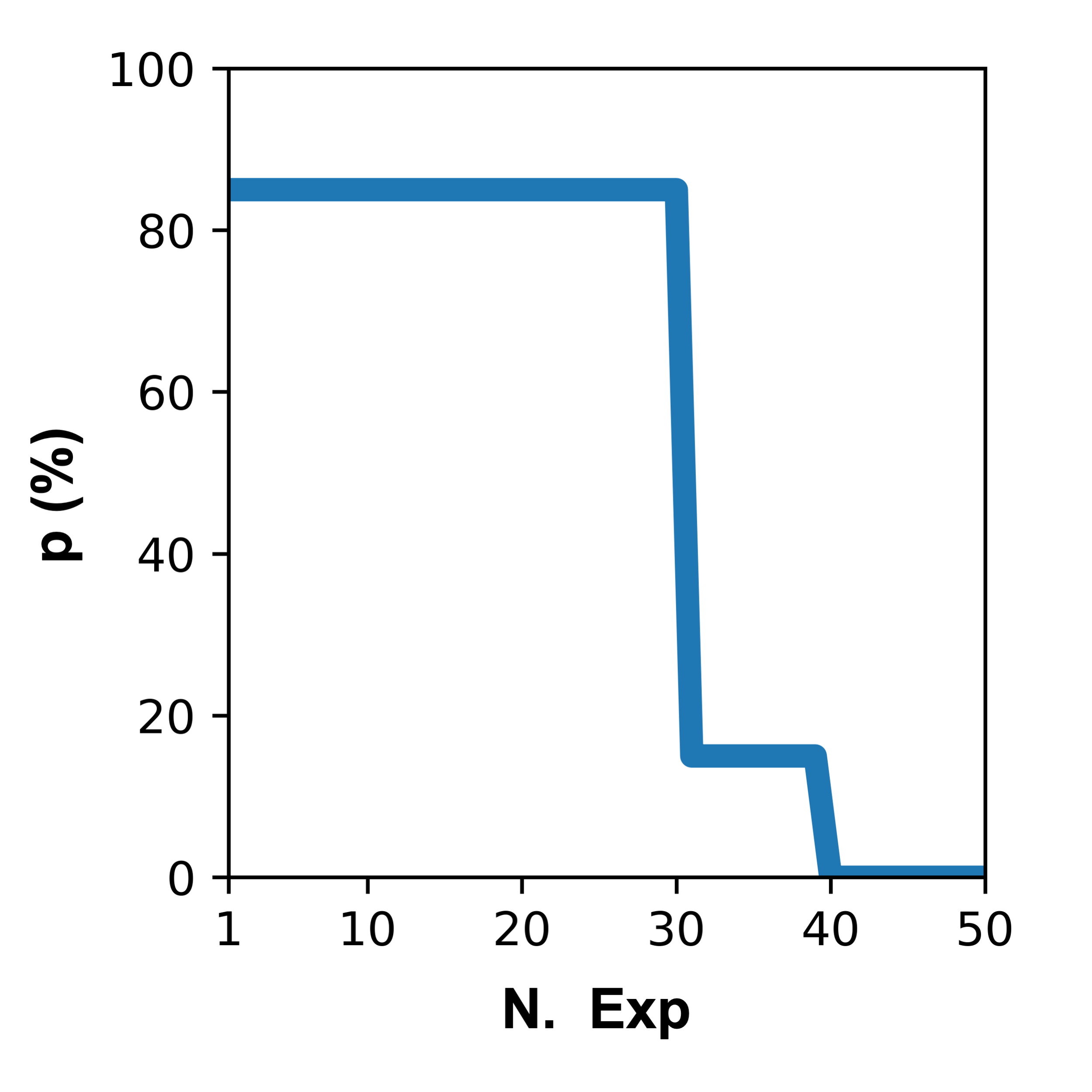}
\caption{\label{fig:pn} Optimized $p(n)$ used for the LLM-EI acquisition function}
\end{figure}

Figure S2 shows an example of typical reasoning used by the LLM to answer survey questions. This example is a randomly selected question from the survey generated for the DA dataset.

\begin{figure}[htb]
\renewcommand\figurename{Fig. S}
\centering
\includegraphics[width=5in,keepaspectratio=True]{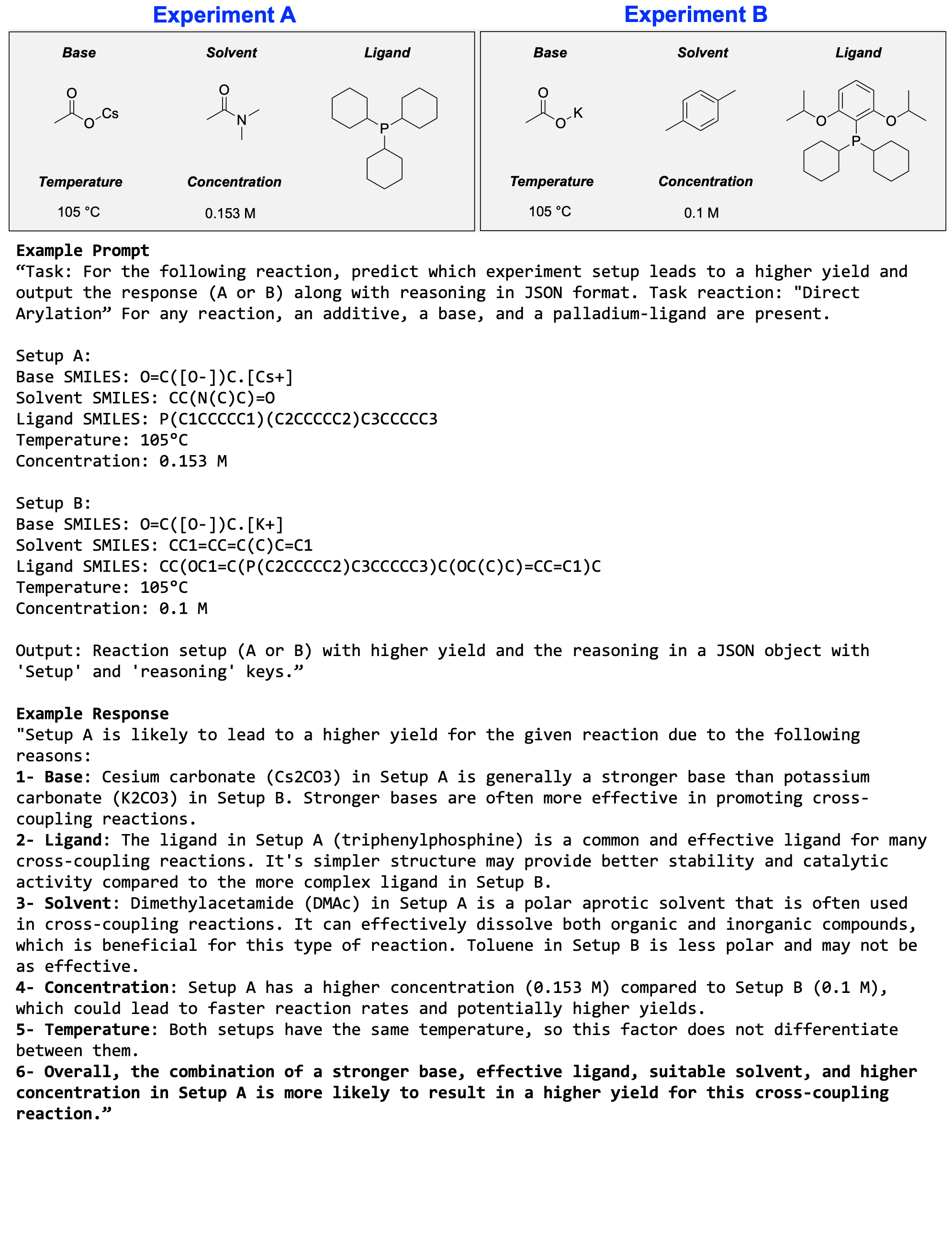}
\caption{\label{fig:reason} Example of LLM reasoning in answering survey}
\end{figure}

Figure S3 provides a comparison between two methods aiming to extract quantitative insights from the LLM. Panel a shows the result of zero-shot LLM regression for the DA dataset, i.e., the LLM was now described the conditions of a single experiment (in the same way that was done for answering the survey) but instead prompted to provided the predicted yield instead of a preference. Panel b shows the utility function $g(x)$ outputs. The utility function outputs show a modest correlation to the true experimentally measured yield, whereas the LLM-predicted yields do not.

\begin{figure}[htb]
\renewcommand\figurename{Fig. S}
\centering
\includegraphics[width=6in,keepaspectratio=True]{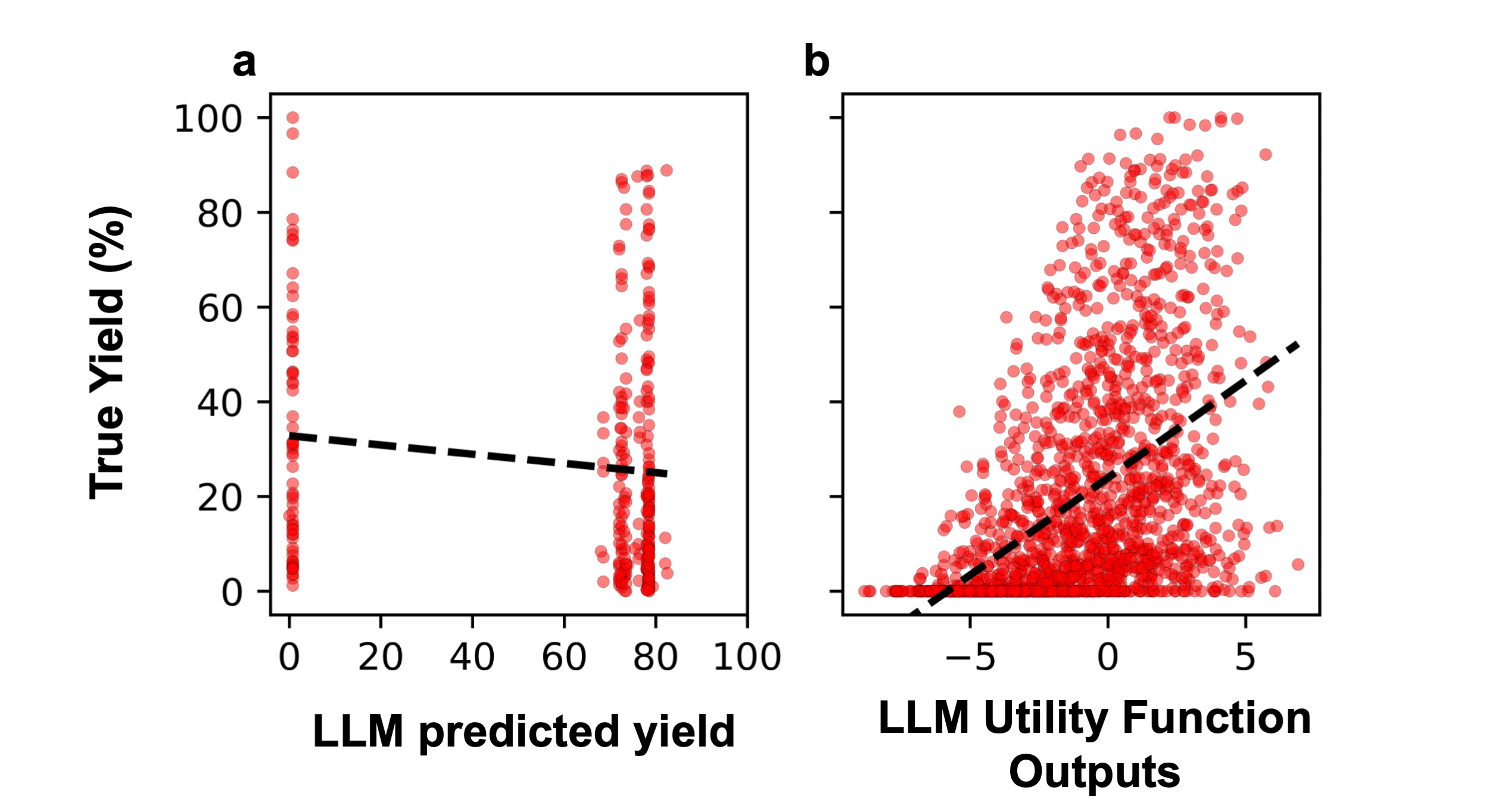}
\caption{\label{fig:pn_comparison} Comparison between simple zero shot regression (panel a) and preference learning on survey (answered by zero-shot LLM) for the DA dataset (panel b)}
\end{figure}

Figure S4 shows the mean (and standard error from n=50 trials) number of experiments needed to run to observe 99\% of the maximum yield for a given dataset and acquisition function. 

\begin{figure}[htb]
\renewcommand\figurename{Fig. S}
\centering
\includegraphics[width=3in,keepaspectratio=True]{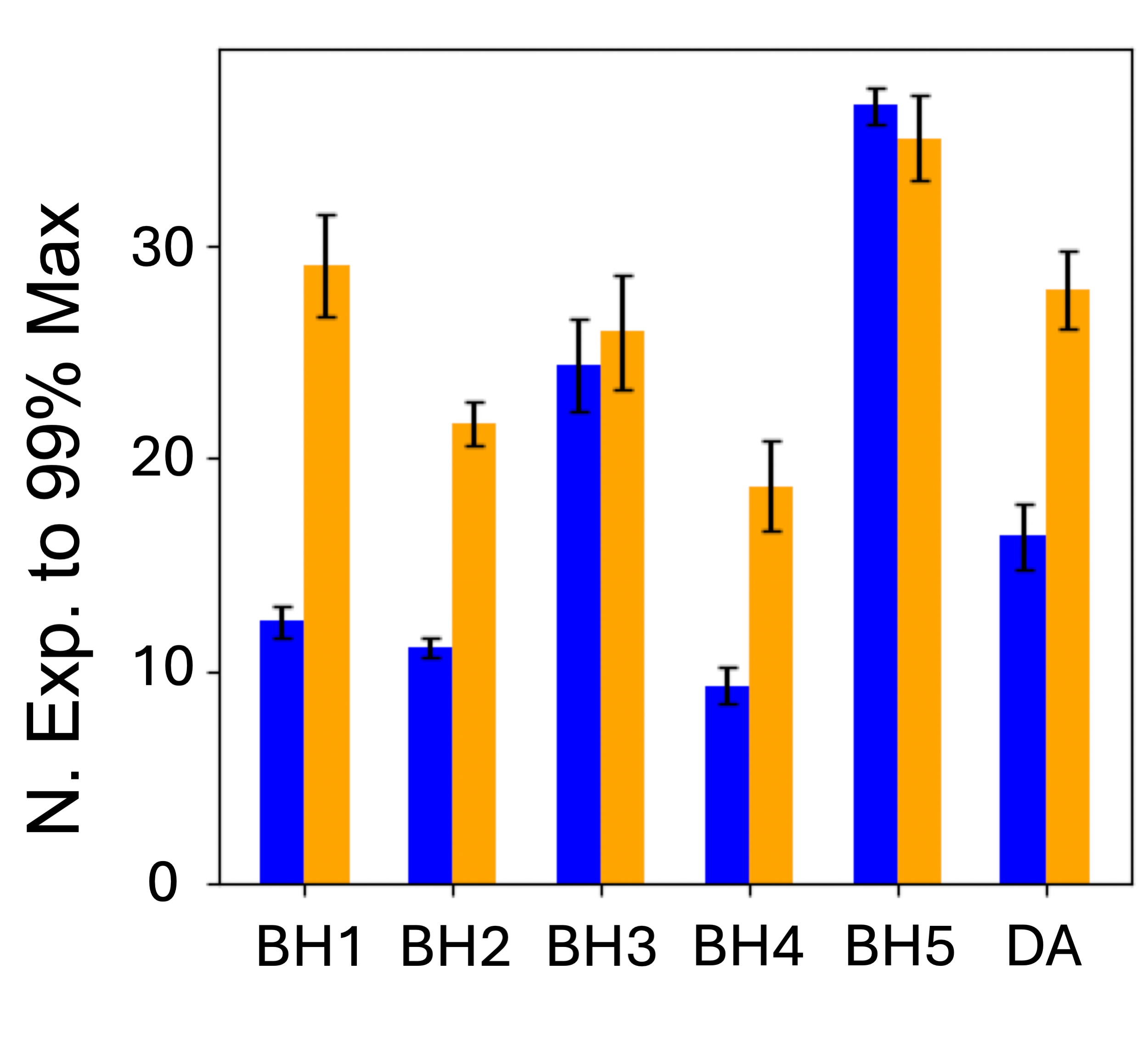}
\caption{\label{fig:n299} Comparison between the LLM-augmented- (blue) and standard expected improvement acquisition function (yellow).}
\end{figure}

\clearpage
\bibliographystyle{unsrtnat}
\bibliography{main} 
\end{document}